\documentclass{article}
\usepackage{iclr2027_conference,times}
\usepackage[T1]{fontenc}
\usepackage{amsmath,amssymb,booktabs,array,graphicx,multirow}
\usepackage{xcolor,colortbl}
\usepackage{hyperref,url}
\pdftrailerid{}
\hypersetup{colorlinks=true,citecolor=blue!55!black,linkcolor=blue!55!black,urlcolor=blue!55!black}

\title{WAM-OPD: Sharpening World Action Models via On-Policy Distillation}
\author{%
\textbf{Panjun Liu, Xiaohan Lei, Shiqi Zhang, Yikun Wang,}\\
\textbf{Yongxin Zhang, Mingyi Hu, Shida Sun, Jiateng Shou,}\\
\textbf{Wengang Zhou, Jiajun Deng\thanks{Corresponding author.}, Zhiwei Xiong}\\
University of Science and Technology of China\\
\texttt{panjun\_liu@mail.ustc.edu.cn, dengjj@ustc.edu.cn}%
}
\iclrfinalcopy
\hypersetup{%
  pdftitle={WAM-OPD: Sharpening World Action Models via On-Policy Distillation},
  pdfauthor={}
}

\begin{document}

\raggedbottom
\maketitle
\lhead{}
\renewcommand{\headrulewidth}{0pt}

\begin{abstract}

Pretrained world action models (WAMs) provide generalist capabilities across diverse robotic manipulation tasks, yet improving target-task performance to an expert level without degrading pretrained skills remains challenging.
To this end, we explore on-policy distillation (OPD) for WAMs and introduce a new framework, namely WAM-OPD. WAM-OPD inherits the advantage of OPD methods that transfer task-specific teacher knowledge under the student’s own induced distribution, rather than directly fitting the student to a narrow task-specific data distribution.
However, in closed-loop manipulation, the observation histories change as the student policy evolves, requiring fresh environment rollouts to remain on-policy.
Directly applying OPD to WAMs entails repeated data collection, which is costly even in simulation and often impractical on real robots.
To avoid repeated environment rollouts during distillation, we introduce prefix-weighted trajectory replay (PWTR). PWTR uses a fixed trajectory pool composed primarily of initial-student rollouts, supplemented with task-specific teacher rollouts to broaden trajectory coverage. For each trajectory replayed from this pool, PWTR conditions the current policy on successive stored histories to generate fresh denoising paths, along which the task-specific teacher provides supervision. Although these denoising paths are refreshed as the policy evolves, the replayed environment trajectories remain fixed. PWTR therefore reweights per-decision distillation losses using proxy importance weights derived from path scores accumulated over the trajectory prefix preceding each decision to mitigate the resulting shift in the history distribution.
Simulated and real-world experiments demonstrate task adaptation without additional environment interaction during distillation. Across both settings, WAM-OPD improves target-task performance while retaining near-initial performance on tasks excluded from adaptation.

\end{abstract}

\vspace{-2mm}
\section{Introduction}
\label{sec:intro}
\begin{figure*}[t]
    \centering
    \includegraphics[width=\textwidth]{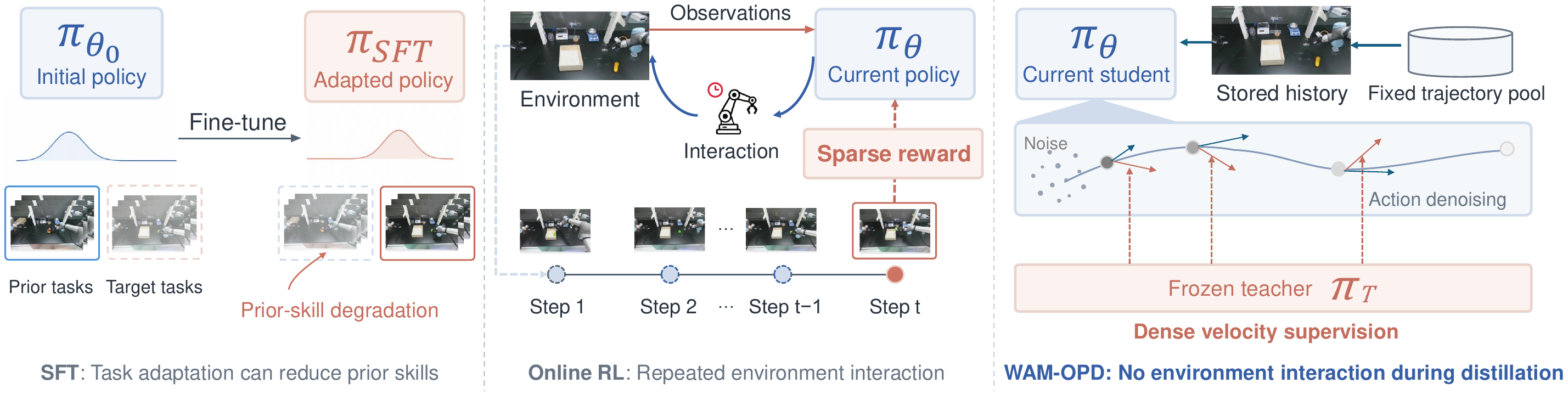}
    \vspace{-8mm}
    \caption{\textbf{Task adaptation strategies for pretrained WAMs.}
\textbf{Left:} SFT specializes pretrained WAMs for target tasks but risks forgetting pretrained skills.
\textbf{Middle:} Online RL couples policy optimization with repeated environment interaction, while sparse rewards can further reduce sample efficiency.
\textbf{Right:} WAM-OPD uses PWTR to replay observation histories from a fixed trajectory pool. A frozen teacher provides dense velocity supervision along fresh student denoising paths, enabling adaptation without additional environment interaction during distillation.}
    \vspace{-4mm}
    \label{fig:teaser}
\end{figure*}

\vspace{-2mm}
Pretrained world action models (WAMs) provide a promising foundation for generalist robotic manipulation by capturing broad, reusable behaviors~\citep{ye2026dreamzero,yuan2026fastwam,li2026lingbotva,li2025uva,zhu2025uwm,kim2026cosmospolicy,agibot2026geact}. Practical deployment, however, often demands higher reliability on frequently executed tasks, since even occasional failures can lead to recurring workflow interruptions and human intervention. This motivates post-training WAMs to achieve expert-level performance on selected tasks while retaining their broader capabilities. Balancing these objectives remains challenging with existing approaches. Supervised fine-tuning (SFT) offers a straightforward and computationally efficient route to task specialization using demonstrations~\citep{kim2025oft}. Reinforcement learning (RL) provides an alternative by directly optimizing task rewards through environmental interaction, enabling substantial improvements in task performance~\citep{li2025simplevlarl}. However, task-specific adaptation can degrade pretrained capabilities under both SFT~\citep{kirkpatrick2017ewc,luo2023forgetting,li2025simplevlarl,zhong2026vlaopd} and RL~\citep{pmlr-v235-wolczyk24a}. Online RL additionally requires repeated environment rollouts~\citep{li2025simplevlarl,qian2026wamrl}, which incur substantial execution and reset costs and must accommodate safety constraints on real robots. Sparse rewards can further increase interaction burden by reducing sample efficiency~\citep{andrychowicz2017her,tan2025riptvla,li2025simplevlarl}. These limitations motivate post-training strategies that jointly address target-task proficiency, skill retention, and interaction efficiency. Figure~\ref{fig:teaser} contrasts these task adaptation strategies.

Recent advances in on-policy distillation (OPD) for large language models~\citep{shenfeld2026self} and visual generation~\citep{li2026diffusionopd,fang2026flowopd} have shown promise in integrating task-specific expertise while mitigating forgetting.
Motivated by these results, we present an early exploration of OPD for WAM adaptation and introduce WAM-OPD.
Particularly, on-policy distillation in WAMs involves two levels of sampling: the student’s denoising paths and the observation histories used as conditioning inputs.
For a given history, the current student generates fresh denoising paths, while a frozen task-specific teacher provides velocity targets at sampled intermediate states.
Generating such denoising paths and obtaining teacher supervision require only model inference.
The observation histories, however, are induced by closed-loop execution of the student’s policy, and the distribution evolves as the policy updates.
Direct online OPD thus requires repeated environment rollouts to collect histories under the updated policy, making it costly even in simulation and often impractical on real robots.

To avoid repeated environment rollouts during distillation, we introduce prefix-weighted trajectory replay (PWTR). PWTR uses a fixed trajectory pool composed primarily of initial-student rollouts, supplemented with task-specific teacher rollouts to broaden trajectory coverage. The initial-student rollouts cover histories typical of the policy at the start of distillation, while teacher rollouts broaden coverage of histories the student may encounter later as it acquires task-specific expertise. For each trajectory replayed from this pool, PWTR conditions the current policy on successive stored histories to generate fresh denoising paths, along which the task-specific teacher provides dense supervision.
Although these denoising paths are refreshed as the policy evolves, the replayed trajectories remain fixed. PWTR therefore reweights per-decision distillation losses to mitigate the resulting distribution shift. PWTR accumulates denoising-path scores under a student snapshot and the fixed initial-student and task-specific teacher references over preceding sampled decision points. These cumulative scores yield proxy importance weights inspired by a change of measure~\citep{precup2000offpolicy}, placing greater emphasis on histories more compatible with the student snapshot relative to the fixed references. Together with fresh denoising paths generated by the current policy, this reweighting provides an approximation to the online OPD objective without further environment interaction during distillation.

Experiments in simulation~\citep{chen2025robotwin} and on real robots demonstrate that WAM-OPD enables task adaptation without additional environment interaction during distillation. Across both settings, WAM-OPD improves target-task performance over the pretrained base model while retaining near-initial performance on tasks excluded from adaptation.
Our contributions are threefold:
\begin{itemize}
    \item To the best of our knowledge, we are the first to explore OPD for pretrained WAMs and introduce WAM-OPD, which transfers task-specific teacher knowledge through dense supervision along denoising paths generated by the current student policy.

    \item We propose PWTR, combining a fixed trajectory pool with prefix-based proxy weighting to mitigate history distribution shift without environment interaction during distillation.

    \item Simulated and real-world experiments show that WAM-OPD improves target-task performance over the pretrained base model while retaining near-initial performance on tasks excluded from adaptation.
\end{itemize}

\vspace{-2mm}
\section{Related Work}

\vspace{-2mm}
\subsection{World Action Models and Task Adaptation}
\vspace{-2mm}
Pretrained WAMs integrate predictive visual modeling with action generation, providing reusable priors for generalist robotic manipulation~\citep{ye2026dreamzero,li2026lingbotva,kim2026cosmospolicy}. Their post-training objectives range from improving control to accelerating inference. WAM-RL jointly improves world and action models through online video fine-tuning and reinforcement learning with reconstruction-based rewards~\citep{qian2026wamrl}. Flash-WAM uses consistency distillation to reduce the denoising steps required for video and action generation~\citep{akbari2026flashwam}. Our work focuses on teacher-guided task adaptation through on-policy distillation, improving target-task performance while retaining pretrained skills.

\vspace{-2mm}
\subsection{On-Policy Distillation}
\vspace{-2mm}
OPD provides teacher supervision on samples generated by the current student. ImitKD and GKD apply this principle to autoregressive generation to reduce the mismatch between training and inference~\citep{lin2020imitkd,agarwal2024gkd}. Flow-OPD and DiffusionOPD extend it to continuous generation, transferring task-specific expertise through teacher feedback along student-generated denoising paths~\citep{fang2026flowopd,li2026diffusionopd}.
In closed-loop manipulation, however, the student determines the observations that condition subsequent predictions. DAgger addresses learner-induced distribution shift through repeated rollouts and expert annotation~\citep{ross2011dagger}. VLA-OPD provides dense teacher guidance on student-visited robot states, while retaining online environment rollouts during training~\citep{zhong2026vlaopd}. Thus, richer supervision alone does not remove the cost of acquiring observation histories consistent with the evolving policy. Our work applies OPD to pretrained WAMs through PWTR, which combines a fixed trajectory pool with proxy prefix weighting to enable adaptation without additional environment interaction during distillation.

\vspace{-2mm}
\subsection{Efficient Distillation and Data Reuse}
\vspace{-2mm}
Data reuse reduces different costs depending on what is retained. Lightning OPD caches responses from an SFT-initialized student and their teacher log-probabilities, eliminating the need for a live teacher server during distillation~\citep{wu2026lightningopd}. ReOPD removes tool-environment interaction by replaying teacher histories for language-agent distillation, with greater supervision weight on earlier steps~\citep{liao2026reopd}. For robotic adaptation, WAM-OPD uses PWTR with a fixed trajectory pool composed primarily of initial-student rollouts, supplemented with task-specific teacher rollouts. Conditioned on stored observation histories, the current student generates fresh denoising paths along which frozen teachers provide dense supervision.
Motivated by off-policy importance sampling and V-trace~\citep{precup2000offpolicy,espeholt2018impala}, PWTR derives proxy prefix weights from accumulated denoising-path scores under a student snapshot and fixed references. These weights are refreshed as the student evolves, adjusting the contribution of replayed histories while keeping the trajectory pool fixed.

\vspace{-2mm}
\section{Method}
\label{sec:method}
\begin{figure*}[t]
    \centering
    \includegraphics[width=\textwidth]{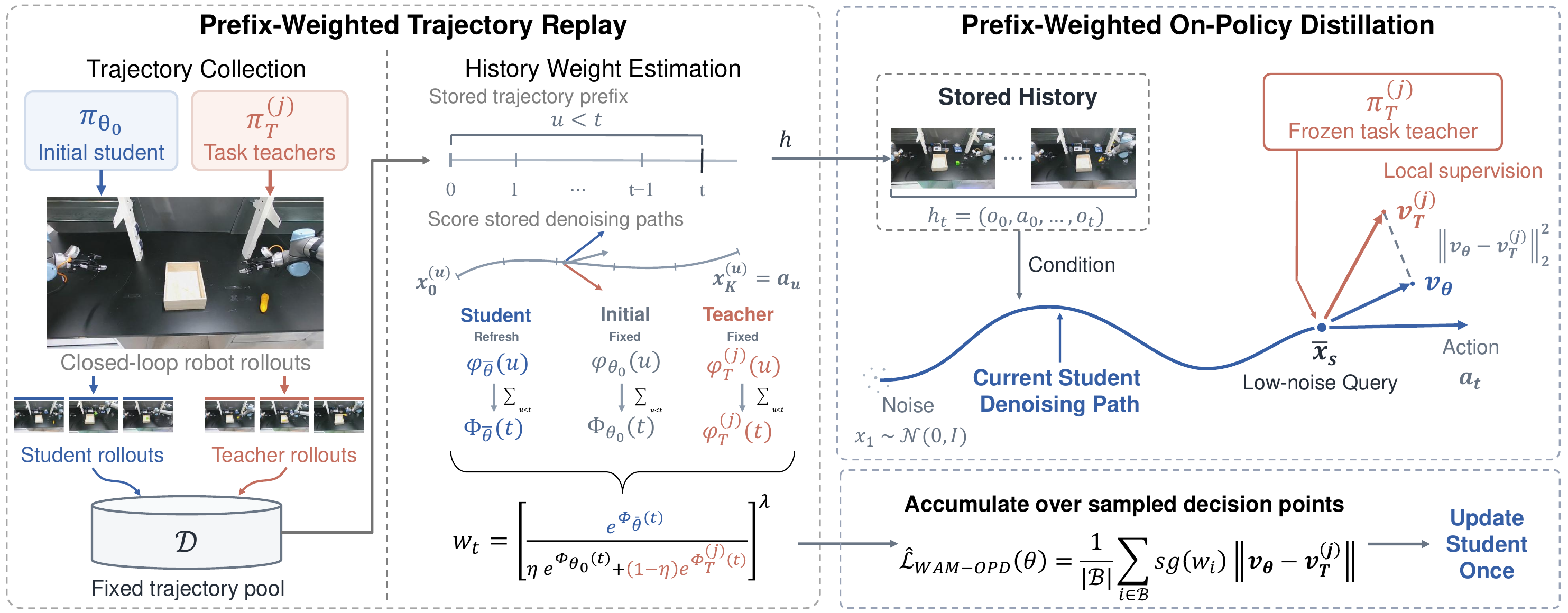}
    \vspace{-8mm}
\caption{\textbf{Overview of WAM-OPD.}
\textbf{Left: Prefix-weighted trajectory replay (PWTR).}
A fixed pool $\mathcal D$ contains primarily initial-student rollouts, supplemented by task-specific teacher rollouts. Stored denoising paths are scored by a student snapshot and the fixed initial-student and teacher references. The cumulative prefix compatibility scores $\Phi_{\bar\theta}(i)$, $\Phi_{\theta_0}(i)$, and $\Phi_T^{(j)}(i)$ yield proxy weights $w_i$.
\textbf{Right: On-policy distillation.}
Given a stored history $h_i$, the current student generates a fresh denoising path. A frozen teacher supervises the student's velocity at a detached low-noise query state $\bar{\mathbf x}_s$ on this path. Prefix-weighted losses are averaged over the sampled decision points of one trajectory, with gradients accumulated for one student update. Here $\operatorname{sg}$ denotes stop-gradient and $\mathcal B$ indexes the trajectory's sampled decision points.}
\vspace{-4mm}
    \label{fig:method}
\end{figure*}

\vspace{-2mm}
WAM-OPD combines prefix-weighted trajectory replay (PWTR) with prefix-weighted on-policy distillation to adapt pretrained WAMs, as illustrated in Figure~\ref{fig:method}. It constructs a fixed pool $\mathcal D$ primarily from rollouts of the initial student $\pi_{\theta_0}$, supplemented with rollouts from task-specific teachers $\pi_T^{(j)}$. At each update, it samples a trajectory from $\mathcal D$ and processes its sampled decision points in temporal order. At sampled decision point $i$, it accumulates scores of stored denoising paths under a student snapshot and the fixed references over preceding sampled points in the same trajectory to obtain prefix scores $\Phi_\nu(i)$. These scores determine the proxy weight $w_i$ to mitigate history distribution mismatch. Conditioned on the stored history $h_i$, the current student $\pi_\theta$ generates a fresh denoising path. At a detached low-noise state $\bar{\mathbf x}_s$, WAM-OPD uses velocity targets $\mathbf v_T^{(j)}$ from the frozen teacher to supervise the student field $\mathbf v_\theta$ and weights the loss by $w_i$. It averages these losses and accumulates gradients over the sampled decision points before updating the student, without further environment interaction.

We first formulate the online OPD objective for WAMs in Section~\ref{sec:opd}. We then introduce PWTR, including the fixed trajectory pool and prefix weight construction, in Section~\ref{sec:pwtr}, followed by prefix-weighted on-policy distillation in Section~\ref{sec:weighted_opd}.

\subsection{On-Policy Distillation Objective for WAMs}
\label{sec:opd}
\vspace{-2mm}
\paragraph{Diffusion-based OPD.}
In a flow-based formulation of diffusion-based OPD~\citep{li2026diffusionopd,fang2026flowopd}, denoising paths are obtained by integrating the current student's velocity field $\mathbf{v}_\theta$ from noise to data, while the frozen teacher's velocity field $\mathbf{v}_T$ provides supervision at intermediate states. For a conditioning input $c$ drawn from a fixed distribution $p(c)$, let $q_\theta(\mathbf{x}_s\mid c,s)$ denote the distribution of student-generated states at denoising time $s$. A corresponding velocity-matching objective is
\begingroup
\setlength{\abovedisplayskip}{3pt plus 1pt minus 2pt}
\setlength{\belowdisplayskip}{3pt plus 1pt minus 2pt}
\setlength{\abovedisplayshortskip}{3pt plus 1pt minus 2pt}
\setlength{\belowdisplayshortskip}{3pt plus 1pt minus 2pt}
\begin{equation}
\label{eq:opd_diff}
\begin{aligned}
\mathcal{L}_{\mathrm{OPD}}(\theta)
&=\mathbb{E}_{c\sim p(c)}
\mathbb{E}_{s\sim p_s}
\mathbb{E}_{\mathbf{x}_s\sim q_\theta(\cdot\mid c,s)}
\Big[
\big\|\mathbf{v}_\theta(\mathbf{x}_s,s\mid c)
-\mathbf{v}_T(\mathbf{x}_s,s\mid c)\big\|_2^2
\Big].
\end{aligned}
\end{equation}
\endgroup
Here, $s=1$ and $s=0$ correspond to noise and data, respectively, and $p_s$ specifies the query-time distribution. The on-policy component lies in sampling intermediate states from $q_\theta$, which evolves with the student, while the conditioning distribution $p(c)$ remains fixed. Conditioning inputs can therefore be reused throughout distillation, with fresh student-generated denoising paths and teacher supervision obtained through model inference alone.

\paragraph{OPD in closed-loop WAMs.}
Closed-loop WAMs introduce policy dependence into the conditioning distribution itself. At each replanning decision $t$, the student generates an action chunk $\mathbf a_t$ conditioned on the history $h_t$. Here, $h_t$ denotes the interaction history available at decision $t$; the policy uses the conditioning inputs extracted from this history. Executing part of this chunk changes the environment and produces subsequent observations. Let $d_\theta$ denote the distribution of histories encountered across replanning decisions during closed-loop execution of the current student policy. Because the student's preceding actions influence subsequent observations and thus the policy inputs, $d_\theta$ evolves with the policy. Replacing the fixed conditioning distribution in Eq.~\eqref{eq:opd_diff} with $d_\theta$ yields the online WAM OPD objective
\begingroup
\setlength{\abovedisplayskip}{3pt plus 1pt minus 2pt}
\setlength{\belowdisplayskip}{3pt plus 1pt minus 2pt}
\setlength{\abovedisplayshortskip}{3pt plus 1pt minus 2pt}
\setlength{\belowdisplayshortskip}{3pt plus 1pt minus 2pt}
\begin{equation}
\label{eq:opd_wam}
\begin{aligned}
\mathcal{L}^{\mathrm{WAM}}_{\mathrm{OPD}}(\theta)
&=\mathbb{E}_{h\sim d_\theta}
\mathbb{E}_{s\sim p_s}
\mathbb{E}_{\mathbf{x}_s\sim q_\theta(\cdot\mid h,s)}
\Big[
\big\|\mathbf{v}_\theta(\mathbf{x}_s,s\mid h)
-\mathbf{v}_T(\mathbf{x}_s,s\mid h)\big\|_2^2
\Big].
\end{aligned}
\end{equation}
\endgroup
Here, $\mathbf{v}_\theta$ and $\mathbf{v}_T$ denote the student and frozen teacher action velocity fields, respectively. Unlike Eq.~\eqref{eq:opd_diff}, the outer expectation follows $d_\theta$, the distribution of histories induced by closed-loop execution of the current student. On-policy sampling therefore involves both histories drawn from $d_\theta$ and denoising states drawn from $q_\theta(\cdot\mid h,s)$. Given a stored history, the current student can regenerate fresh denoising states through model inference, but the history itself need not follow the current $d_\theta$. Obtaining fresh histories from $d_\theta$ requires closed-loop execution, so directly realizing online OPD entails repeated environment rollouts~\citep{zhong2026vlaopd}, which are costly in simulation and often impractical on real robots. To approximate this objective using a fixed trajectory pool without further environment interaction during distillation, WAM-OPD uses PWTR, described in Section~\ref{sec:pwtr}.

\subsection{Prefix-Weighted Trajectory Replay}
\label{sec:pwtr}
\label{sec:collection}
\label{sec:sampling}

\paragraph{Fixed trajectory pool.}
As illustrated in Figure~\ref{fig:method}, WAM-OPD constructs a fixed trajectory pool $\mathcal{D}=\mathcal{D}_S\cup\mathcal{D}_T$ before distillation. The pool consists primarily of initial-student trajectories $\mathcal{D}_S$, collected by executing $\pi_{\theta_0}$ across all target tasks, supplemented with teacher trajectories $\mathcal{D}_T$ collected by executing each frozen teacher $\pi_T^{(j)}$ on its assigned task. Initial-student rollouts provide histories encountered before adaptation, including unsuccessful executions, while teacher rollouts broaden coverage with histories from task-specific policies. The pool remains fixed throughout distillation.

At each replanning decision $t$, the collecting policy generates an action chunk $\mathbf{a}_t$ of horizon $H$ and executes its first $\ell_t\leq H$ actions before receiving the next observation. Each stored record contains the history $h_t$, from which the collecting policy's conditioning inputs are extracted, together with the sampled action chunk $\mathbf{a}_t$, the executed length $\ell_t$, and the complete denoising path that produced $\mathbf{a}_t$. Each trajectory also carries task and source labels: the task identifies the supervising teacher, and the source distinguishes initial-student from teacher rollouts. The executed length $\ell_t$ determines which action coordinates are included in path scoring.

Stored denoising paths are used to estimate prefix weights. We use equally spaced temporal subsampling with $n=16$, preserving the order of the sampled decision points within each trajectory. The path score at a sampled decision point contributes only to the weights of later sampled points in the same trajectory. Stored action chunks are not used as regression targets.

\paragraph{Reweighting replayed histories.}
PWTR reweights the distillation loss to account for the changing history distribution as the student evolves. During training, it selects the initial-student or teacher source with probabilities $\eta$ and $1-\eta$, respectively, and samples a trajectory from that source. At the population level, this sampling scheme defines the reference mixture
$p_{\mathrm{mix}}=\eta d_{\theta_0}+(1-\eta)d_T$,
where $d_{\theta_0}$ and $d_T$ are the initial-student and task-specific teacher history distributions under a common temporal sampling convention. When $p_{\mathrm{mix}}$ covers the support of $d_\theta$, weighting by $d_\theta(h)/p_{\mathrm{mix}}(h)$ recovers the online expectation over histories~\citep{precup2000offpolicy,sutton2018rl}.

PWTR constructs proxy weights for replayed histories through their associated trajectory prefixes. Under the assumptions in Appendix~\ref{app:history_reweighting}, shared initialization and environment-transition factors cancel in the full-prefix density ratio at a fixed original decision index, leaving a ratio determined by preceding action probabilities. This structure motivates weighting each history using the actions that precede it. Since the flow sampler does not directly provide these probabilities, PWTR scores stored denoising paths by their compatibility with each policy's velocity field and accumulates these scores over preceding sampled decision points to construct the weights.

\paragraph{Denoising path scoring.}
For a stored denoising path at replanning decision $u$, let
$\{\mathbf{x}_{s_m}\}_{m=0}^{K}$
denote its states at noise levels $s_0>\cdots>s_K$, with
$\Delta_m=s_{m+1}-s_m<0$.
PWTR evaluates the path under a scoring policy $\nu$, corresponding to the student snapshot, initial student, or task-specific teacher:
\begingroup
\setlength{\abovedisplayskip}{3pt plus 1pt minus 2pt}
\setlength{\belowdisplayskip}{3pt plus 1pt minus 2pt}
\setlength{\abovedisplayshortskip}{3pt plus 1pt minus 2pt}
\setlength{\belowdisplayshortskip}{3pt plus 1pt minus 2pt}
\begin{equation}
\label{eq:path_score}
\begin{aligned}
\varphi_\nu(u)
&=-\sum_{m=0}^{K-1}\frac{1}{2\sigma_m^2 d_u}
\left\|
P_u\left[
\mathbf{x}_{s_{m+1}}-\mathbf{x}_{s_m}
-\Delta_m\mathbf{v}_\nu(\mathbf{x}_{s_m},s_m\mid h_u)
\right]
\right\|_2^2.
\end{aligned}
\end{equation}
\endgroup
The residual compares each recorded denoising increment with the increment predicted by policy $\nu$ at the same stored state and history $h_u$. The projection $P_u$ selects the $d_u$ coordinates corresponding to the $\ell_u$ executed actions, restricting scoring to the portion of the chunk applied to the environment. Division by $d_u$ normalizes for the number of scored coordinates, while
$\sigma_m=\max\{\sigma_{\min},\kappa\sqrt{-\Delta_m}\}$
provides an auxiliary scoring scale, with $\sigma_{\min}>0$ and $\kappa>0$. Higher scores indicate closer agreement between the policy's velocity field and the stored path. The resulting $\varphi_\nu(u)$ serves as a path-compatibility proxy for prefix weighting, rather than an exact action log-likelihood.

\paragraph{Prefix weight construction.}
Let $t_{\tau,k}$ denote the original decision index of the $k$-th sampled decision point in trajectory $\tau$. For $i=(\tau,k)$, we define the prefix score as
$\Phi_\nu(i)=\sum_{r<k}\varphi_\nu(t_{\tau,r})$,
where the sum is zero for the first sampled point. The current point is excluded because its history is available before its action is generated. Following the mixture structure of the full-prefix density ratio, PWTR compares a student snapshot $\bar\theta$ with the initial-student and task-specific teacher references, as illustrated in Figure~\ref{fig:method}:
\begingroup
\setlength{\abovedisplayskip}{3pt plus 1pt minus 2pt}
\setlength{\belowdisplayskip}{3pt plus 1pt minus 2pt}
\setlength{\abovedisplayshortskip}{3pt plus 1pt minus 2pt}
\setlength{\belowdisplayshortskip}{3pt plus 1pt minus 2pt}
\begin{equation}
\label{eq:cum_score}
S_i
=\lambda\left[
\Phi_{\bar\theta}(i)
-\log\left(
\eta e^{\Phi_{\theta_0}(i)}
+(1-\eta)e^{\Phi_T^{(j)}(i)}
\right)
\right].
\end{equation}
\endgroup
Here, $\lambda\geq 0$ controls the reweighting strength, and $\exp(S_i)$ defines an unnormalized proxy weight. Larger weights favor replayed records whose preceding denoising paths are more compatible with the student snapshot relative to the fixed references. Both fixed references score every stored path to form the mixture denominator.

Within each collection condition, PWTR normalizes $\exp(S_i)$ to unit mean under the empirical training measure, clips the normalized weights at threshold $C$, and renormalizes them under the same measure to obtain $w_i$ (Appendix, Eq.~\eqref{eq:weights}). This measure follows the trajectory-sampling probabilities and the averaging of per-decision losses within each trajectory. Clipping limits the influence of extreme scores, while renormalization maintains unit mean within each condition. Setting $\lambda=0$ yields $w_i=1$, recovering unweighted mixed-source distillation.

The fixed-reference scores are computed once for all stored paths and cached. Every $R$ student updates, PWTR refreshes the snapshot $\bar\theta$, evaluates its velocity field at the stored denoising states, and recomputes the prefix weights. The weights remain fixed between refreshes.

\subsection{Prefix-Weighted On-Policy Distillation}
\label{sec:weighted_opd}

\paragraph{Local velocity supervision.}
Each sampled trajectory is processed in temporal order. Let $\mathcal B=\mathcal B_\tau$ index the sampled decision points of trajectory $\tau$. Each record $i$ provides the stored history $h_i$, a task index $j_i$, and a cached prefix weight $w_i$, where $h_i=h_{t_{\tau,k}}$ is the interaction history available at that decision. Conditioned on $h_i$, the current student generates a fresh denoising path from Gaussian noise. At a low-noise query time $s_i\sim p_s$, the sampled state is detached as
$\bar{\mathbf{x}}_i=\operatorname{sg}(\mathbf{x}_{s_i})$,
where $\operatorname{sg}$ denotes stop-gradient. As illustrated in Figure~\ref{fig:method}, the student and frozen task-specific teacher are evaluated at the same detached state $\bar{\mathbf{x}}_i$. The teacher's action velocity provides a local supervision target for the student along this freshly generated denoising path.

\paragraph{Trajectory-level student update.}
We average the prefix-weighted velocity-matching losses over the sampled decision points of each trajectory:
\begingroup
\setlength{\abovedisplayskip}{3pt plus 1pt minus 2pt}
\setlength{\belowdisplayskip}{3pt plus 1pt minus 2pt}
\setlength{\abovedisplayshortskip}{3pt plus 1pt minus 2pt}
\setlength{\belowdisplayshortskip}{3pt plus 1pt minus 2pt}
\begin{equation}
\label{eq:final_loss}
\begin{aligned}
\widehat{\mathcal{L}}_{\mathrm{WAM\text{-}OPD}}(\theta)
&=\frac{1}{|\mathcal{B}|}
\sum_{i\in\mathcal{B}}\operatorname{sg}(w_i)
\cdot
\left\|
\mathbf{v}_\theta(\bar{\mathbf{x}}_i,s_i\mid h_i)
-\mathbf{v}_T^{(j_i)}(\bar{\mathbf{x}}_i,s_i\mid h_i)
\right\|_2^2.
\end{aligned}
\end{equation}
\endgroup
Query states, prefix weights, and teacher targets are treated as constants, so gradients propagate only through the current student's velocity predictions. We accumulate gradients over these points before a single student update. The loss supervises only the action velocity field; the video branch receives no direct distillation loss.
This objective combines teacher supervision on fresh current-student denoising paths with proxy prefix weights that mitigate the history distribution mismatch, enabling distillation from the fixed trajectory pool without further environment interaction.

\section{Experiments}
\label{sec:experiments}

We evaluate whether WAM-OPD can improve target-task performance while retaining pretrained skills, analyze the trajectory-pool and weighting choices in PWTR, and examine its applicability to real-world manipulation.
All WAM-OPD distillation updates use a fixed trajectory pool without additional environment interaction.

\subsection{Experimental Setup}
\label{sec:exp_setup}
\label{sec:exp_impl}

\begin{table*}[t]
\centering
\setlength{\belowcaptionskip}{4pt}
\caption{Success rates on RoboTwin.
Both teachers supervise WAM-OPD; underlining marks teaching tasks.
Parentheses show changes in group averages from the initial student.
Bold denotes column maxima excluding Online OPD,
which uses additional environment rollouts during distillation.}
\label{tab:sim_main}

\fontsize{8}{9.5}\selectfont
\renewcommand{\arraystretch}{1.02}

\definecolor{simOurs}{HTML}{F3F7FB}
\definecolor{simTeacher}{HTML}{FFFAF2}
\definecolor{simReference}{HTML}{F7F7F7}
\definecolor{simAccent}{HTML}{486B85}

\newcommand{\simscore}[2]{#1\,{\scriptsize #2}}

\newcommand{\simtablebody}{%
\toprule
\multirow{2}{*}{Method}
& \multicolumn{2}{c}{Target tasks}
& \multicolumn{6}{c}{Retention tasks}
& \multicolumn{2}{c}{Average} \\
\cmidrule(lr){2-3}\cmidrule(lr){4-9}\cmidrule(l){10-11}
& Stapler & Microwave
& Switch & Mug & Laptop & Pillbottle & Dual Bottles & Handover
& Target & Retention \\
\midrule

Initial student
& 67.5 & 45.5
& 68.5 & \textbf{68.5} & \textbf{99.0} & \textbf{99.0}
& \textbf{98.0} & 90.0
& 56.50 & \textbf{87.17} \\

\midrule
\rowcolor{simTeacher}
Stapler teacher
& \underline{\textbf{89.5}} & 33.5
& 52.5 & 66.0 & 87.5 & 97.0 & 95.0 & 56.5
& 61.50 & 75.75 \\

\rowcolor{simTeacher}
Microwave teacher
& 31.0 & \underline{60.5}
& 62.0 & 26.0 & 98.5 & 92.5 & 97.5 & 2.5
& 45.75 & 63.17 \\

\midrule
$\pi$RL
& 0.5 & \textbf{96.0}
& 25.0 & 22.0 & 17.5 & 22.0 & 89.0 & 1.0
& \simscore{48.25}{($-8.25$)}
& \simscore{29.42}{($-57.75$)} \\

STEAM
& 76.0 & 39.0
& 61.5 & 42.5 & 97.5 & 95.5 & 91.5 & 67.5
& \simscore{57.50}{($+1.00$)}
& \simscore{76.00}{($-11.17$)} \\

\rowcolor{simOurs}
\textcolor{simAccent}{\textbf{WAM-OPD (Ours)}}
& 80.5 & 59.0
& \textbf{71.5} & 68.0 & 98.5 & 98.0 & 95.0 & \textbf{90.5}
& \simscore{\textbf{69.75}}{
    \textcolor{simAccent}{\textbf{(+13.25)}}
  }
& \simscore{86.92}{($-0.25$)} \\

\midrule
\rowcolor{simReference}
Online OPD
& 81.5 & 59.0
& 68.0 & 68.5 & 97.0 & 98.0 & 98.5 & 88.0
& \simscore{70.25}{($+13.75$)}
& \simscore{86.33}{($-0.83$)} \\

\bottomrule
}

\newsavebox{\simMeasureBox}
\sbox{\simMeasureBox}{%
  \setlength{\tabcolsep}{0pt}%
  \begin{tabular}{l*{10}{c}}
    \simtablebody
  \end{tabular}%
}
\setlength{\tabcolsep}{
  \dimexpr(\textwidth-\wd\simMeasureBox)/22\relax
}

\begin{tabular}{l*{10}{c}}
  \simtablebody
\end{tabular}
\end{table*}

\paragraph{Tasks and evaluation.}
On RoboTwin 2.0~\citep{chen2025robotwin}, we jointly adapt a single student on Move Stapler Pad and Open Microwave, abbreviated as Stapler and Microwave.
We evaluate skill retention on six tasks excluded from adaptation: Switch, Mug, Laptop, Pillbottle, Dual Bottles, and Handover.
The same adapted student checkpoint is evaluated on both target and retention tasks.
Evaluation uses 100 independent scenes per task and setting, with two runs per scene, yielding 200 trials per task and setting.
All methods share the evaluation scenes, instructions, and inference-noise seeds.
Target and retention averages assign equal weight to their two and six constituent tasks, respectively.
Additional evaluation and interaction-cost details are provided in Appendix~\ref{app:evaluation}.

\paragraph{Compared policies.}
We compare WAM-OPD with the initial student, two task-specific SFT teachers, $\pi$RL~\citep{chen2025pirl}, and STEAM~\citep{liu2026steam} under the same evaluation protocol.
The Stapler and Microwave teachers are independently fine-tuned from the same pretrained checkpoint, evaluated on all eight tasks, and jointly supervise one WAM-OPD student.
Controlled variants are analyzed in Section~\ref{sec:exp_ablation}.

\paragraph{Fixed-pool distillation protocol.}
For each target task, we collect 40 initial-student trajectories: 8 under \texttt{demo\_clean} and 32 under \texttt{demo\_randomized}.
Adding 20 teacher rollouts for each target task yields a fixed pool of 120 trajectories.
These teacher rollouts are distinct from the demonstrations used for teacher SFT.
During distillation, the current student generates fresh denoising paths conditioned on stored histories, with supervision from the corresponding task teacher and prefix weights from PWTR (Section~\ref{sec:pwtr}).
The pool remains fixed, and distillation requires no additional environment interaction.

\paragraph{Implementation details.}
The student and teachers use Fast-WAM~\citep{yuan2026fastwam}, initialized from its RoboTwin checkpoint.
We optimize all student MoT parameters using AdamW at a learning rate of $10^{-6}$ for 96 updates. We use equally spaced temporal subsampling with $n=16$ and average the loss over the sampled decision points of one trajectory per update.
The loss matches student and teacher action velocities at low-noise states, with no direct video supervision.
All evaluated policies use 10 denoising steps to generate 32-step action chunks, executing up to 24 steps before replanning.
Pool composition, backbone details, and additional hyperparameters appear in Appendix~\ref{app:configuration}; prefix-weight normalization and clipping are detailed in Appendix~\ref{app:history_reweighting}.

\subsection{Target-Task Adaptation and Skill Retention}
\label{sec:exp_results}
\label{sec:exp_sim}

Table~\ref{tab:sim_main} reports target-task adaptation and skill retention for the same student policy.
WAM-OPD increases mean target-task success from 56.50\% to 69.75\%, while maintaining 86.92\% mean success on the six retention tasks compared with the initial student's 87.17\%.
This corresponds to a 13.25-percentage-point target-task gain and a 0.25-point decrease in mean retention performance.

\paragraph{Target-task adaptation.}
WAM-OPD improves success from 67.5\% to 80.5\% on Stapler and from 45.5\% to 59.0\% on Microwave, yielding gains of 13.0 and 13.5 percentage points.
WAM-OPD achieves the highest mean target-task success rate among the evaluated SFT and RL baselines, with performance comparable to online OPD (69.75\% vs.\ 70.25\%), without additional environment interaction during distillation.
The SFT specialists exhibit cross-task degradation: the Stapler teacher obtains 33.5\% on Microwave, while the Microwave teacher obtains 31.0\% on Stapler.
WAM-OPD instead improves both target tasks within one student, combining guidance from the two specialists.

\paragraph{Pretrained skill retention.}
The target-task gains are accompanied by near-initial performance on the six evaluated retention tasks.
Both SFT teachers underperform the initial student on all six tasks, with mean success rates of 75.75\% and 63.17\%, compared with 86.92\% for WAM-OPD.
For example, Handover success falls from 90.0\% to 56.5\% for the Stapler teacher and 2.5\% for the Microwave teacher, whereas WAM-OPD achieves 90.5\%.
Together, the target and retention results show that the complete WAM-OPD method improves the adapted skills while retaining near-initial performance on tasks excluded from adaptation.

\subsection{Analysis of Prefix-Weighted Trajectory Replay}
\label{sec:exp_ablation}

\begin{table*}[t]
\centering
\setlength{\belowcaptionskip}{6pt}
\caption{Ablation study of PWTR.
Success rates are reported over 200 cases per task;
M-Shuffle permutes weights among decision points at the same position $k$ in the sampled sequence.}
\label{tab:ablation}
\footnotesize
\setlength{\tabcolsep}{3pt}
\begin{tabular*}{0.90\linewidth}{@{\extracolsep{\fill}}l*{7}{c}@{}}
\toprule
\multirow{2}{*}{Task}
& \multicolumn{2}{c}{Student Pool}
& \multicolumn{2}{c}{Teacher Pool}
& \multicolumn{3}{c}{Mixed Pool} \\
\cmidrule(lr){2-3}
\cmidrule(lr){4-5}
\cmidrule(l){6-8}
& w/o prefix & w/ prefix
& w/o prefix & w/ prefix
& w/o prefix & M-Shuffle & \textbf{w/ prefix (Ours)} \\
\midrule
Stapler
& 75.0 & 77.5 & 74.0 & 72.0
& 77.0 & 74.0 & \textbf{80.5} \\
Microwave
& 57.0 & 58.0 & 54.0 & 55.5
& 57.5 & 57.0 & \textbf{59.0} \\
\midrule
Avg.
& 66.00 & 67.75 & 64.00 & 63.75
& 67.25 & 65.50 & \textbf{69.75} \\
\bottomrule
\end{tabular*}
\end{table*}

\begin{figure}[t]
    \centering
    \includegraphics[width=0.9\linewidth]
    {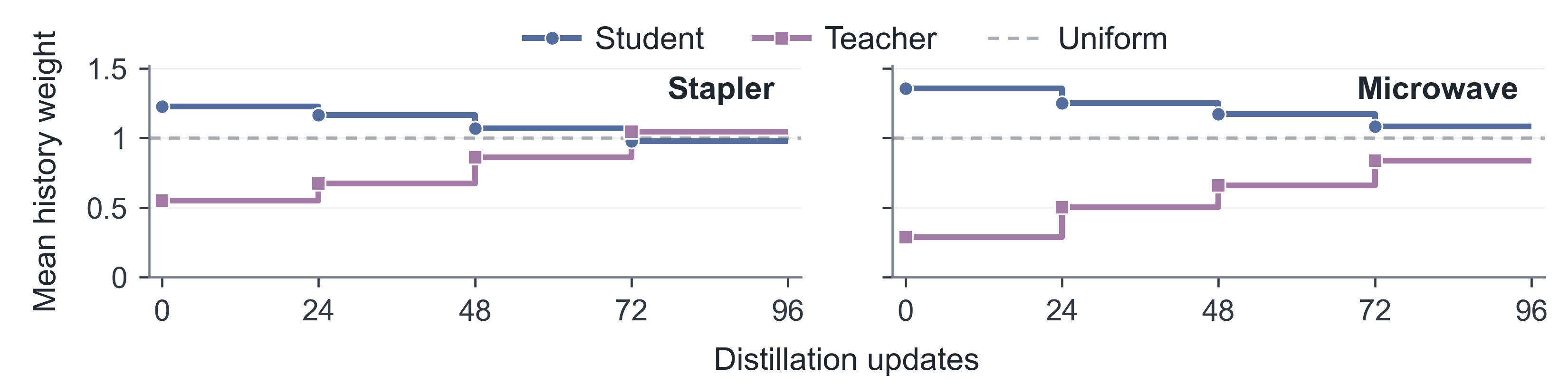}
    \vspace{-4mm}
    \caption{
    \textbf{History weight dynamics during distillation.}
    Mean history weights at sampled decision points are shown
    separately for initial-student and teacher rollouts in the
    mixed trajectory pool.
    The dashed line indicates uniform weighting.
    }
    \vspace{-4mm}
    \label{fig:prefix-weights}
\end{figure}

\paragraph{Trajectory-pool composition.}
The \emph{Student Pool} contains trajectories from the initial student, the \emph{Teacher Pool} contains task-specific teacher rollouts, and the \emph{Mixed Pool} combines both sources.
With prefix weights set to one, the Mixed Pool achieves 67.25\% mean success, compared with 66.00\% for the Student Pool and 64.00\% for the Teacher Pool, as shown in Table~\ref{tab:ablation}.
The Mixed Pool yields the highest success on both tasks with prefix weighting.
These comparisons favor the mixed-pool configuration, in which initial-student trajectories are supplemented with teacher trajectories.

\paragraph{Effect of prefix weighting.}
For each pool, \emph{w/o prefix} assigns unit weight to the distillation loss at each sampled decision point, while \emph{w/ prefix} applies PWTR weights.
As shown in Table~\ref{tab:ablation}, weighting increases mean success from 66.00\% to 67.75\% in the Student Pool and from 67.25\% to 69.75\% in the Mixed Pool, yielding gains of 1.75 and 2.50 percentage points, respectively.
In the Teacher Pool, mean success changes from 64.00\% to 63.75\%, showing no average gain.
Thus, prefix weighting benefits both pools containing initial-student rollouts, with the largest gain in the Mixed Pool.
The Teacher Pool result may reflect limited coverage of student-visited histories: reweighting adjusts the contribution of stored histories but cannot expand their coverage.
Figure~\ref{fig:prefix-weights} also shows how mean history weights evolve for initial-student and teacher rollouts during Mixed Pool distillation.

\paragraph{History--weight correspondence.}
Within the Mixed Pool, \emph{M-Shuffle} randomly permutes PWTR weights among decision points at the same position $k$ in the sampled sequence.
This preserves the weight values at each sampled position while disrupting the correspondence between individual histories and their weights.
Full PWTR achieves 69.75\% mean success versus 65.50\% for M-Shuffle, a gain of 4.25 percentage points.
This result supports the importance of history--weight correspondence.
Together, these ablations favor a mixed trajectory pool with history-specific prefix weighting.

\subsection{Real-World Evaluation}
\label{sec:exp_real}

\paragraph{Platform and tasks.}
The platform comprises two UR5e arms, a head camera providing a global view, and one wrist camera per arm providing local views.
We evaluate Pick-and-Place Corn, Pick-and-Place Block, and Press Button, abbreviated as Corn, Block, and Button.
The first two tasks involve approaching, grasping, transporting, and placing objects, while Button assesses end-effector alignment and contact.
The student is initialized from the Fast-WAM RoboTwin checkpoint and pretrained on real-robot data to obtain a base student adapted to the robot's observation and action interfaces.
Task-specific teachers are initialized from this shared checkpoint, fine-tuned on their respective single-task datasets, and frozen during distillation.

\paragraph{Fixed-pool distillation.}
For each task, trajectories are collected once using the base student and combined with teacher-provided trajectories to form a fixed pool.
WAM-OPD uses supervision from all three task-specific teachers to adapt a single student.
As in simulation, fresh student denoising paths and teacher targets are generated using stored histories, with no additional robot rollouts during distillation.
The real-robot observation and action interfaces are detailed in Appendix~\ref{app:configuration}.

\paragraph{Evaluation protocol.}
Evaluation layouts are independent of those used for training-data collection.
Each method is evaluated on 50 paired layouts per task, with methods tested on the same layouts in interleaved order.
A pick-and-place trial succeeds if the object reaches the designated region and remains stable after release; a button-press trial succeeds if the button is activated.
We report task success rates for the base student, the corresponding task-specific teachers, and the single student adapted by WAM-OPD.

\begin{figure*}[t]
    \centering
    \includegraphics[width=\textwidth]{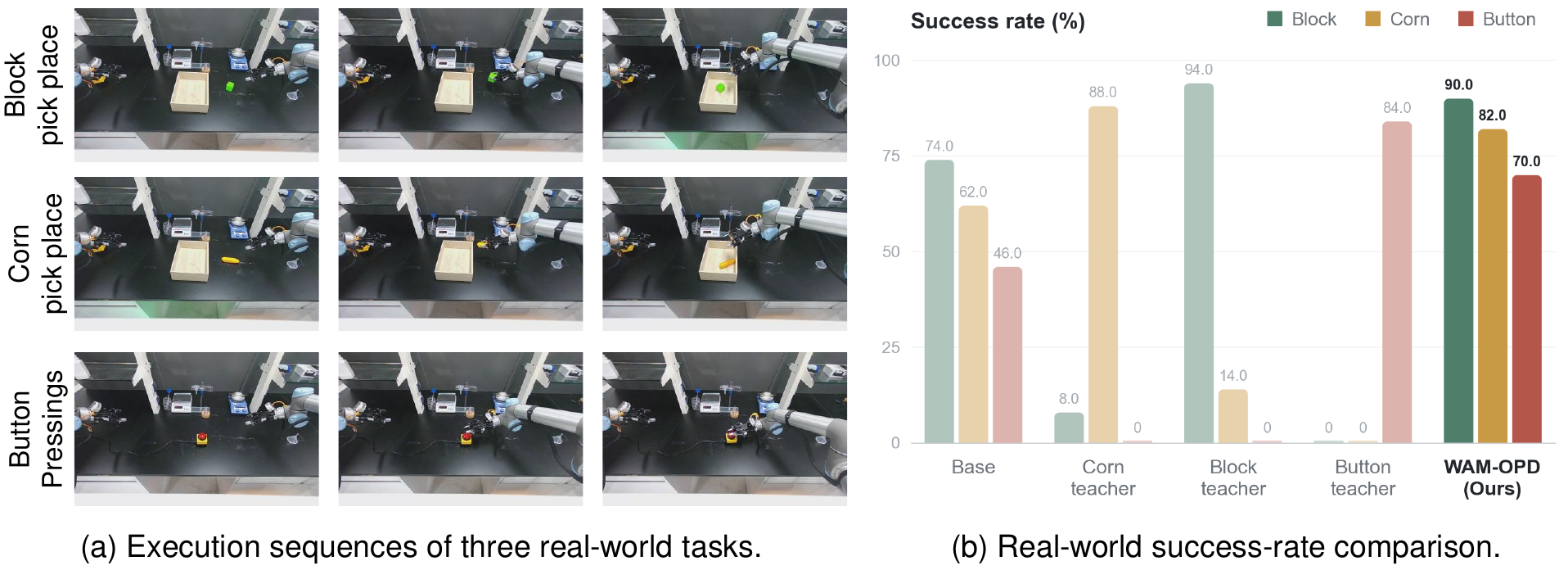}
    \vspace{-6mm}
    \caption{
    Real-world evaluation on Corn, Block, and Button.
    \textbf{Left:} Experimental scenes for the three manipulation tasks.
    \textbf{Right:} Success rates of the base student, the corresponding task-specific teachers, and WAM-OPD.
    WAM-OPD adapts a single student using all three teachers and a fixed trajectory pool, without additional robot rollouts during distillation.
    }
    \vspace{-4mm}
    \label{fig:REAL}
\end{figure*}

\paragraph{Results.}
Figure~\ref{fig:REAL} presents the task scenes and success-rate comparisons.
In this three-task evaluation, WAM-OPD improves upon the base student, supporting target-task adaptation on the real robot within a single policy.
These gains are achieved using the fixed trajectory pool throughout distillation, without additional robot interaction during that stage.

\section{Conclusion}
We presented WAM-OPD, an on-policy distillation framework for adapting pretrained world action models. Experiments in simulation and on real robots show that WAM-OPD improves target-task performance over the pretrained base model while retaining near-initial performance on tasks excluded from adaptation. Through prefix-weighted trajectory replay (PWTR), these gains are achieved using a fixed trajectory pool without additional environment interaction during distillation. These findings support OPD as a promising approach to task specialization with pretrained skill retention. Future work could improve trajectory coverage and proxy prefix-weight estimation to further strengthen adaptation.
\label{main-end}

\bibliography{references}
\bibliographystyle{iclr2027_conference}
\clearpage
\appendix
\section{History-Prefix Reweighting}
\label{app:history_reweighting}

\subsection{Ideal History Ratios}

We consider complete interaction histories conditional on a task and collection condition $g$. Histories are indexed by original environment replanning decisions, with a common absorbing continuation after termination used for this ideal fixed-index analysis. Let $a_u^{\mathrm{exec}}$ denote the action prefix executed between decisions $u$ and $u+1$. For a stochastic policy $\nu$ with executed-action density $\pi_\nu^{\mathrm{exec}}$, a shared initial distribution $p_0$, and a shared environment kernel $P$, the history density at decision $t$ is
\begin{equation}
d_{\nu,t}(h_t)
=p_0(o_0)\prod_{u<t}
\pi_\nu^{\mathrm{exec}}(a_u^{\mathrm{exec}}\mid h_u)
P(o_{u+1}\mid h_u,a_u^{\mathrm{exec}}).
\tag{6}
\end{equation}
The task instruction is included in the conditioning, which is otherwise suppressed in the notation. The product includes every preceding decision. Different initial-state distributions or environment dynamics would introduce additional factors into the density ratio.

The collection source is selected once per trajectory. Writing $\eta_g$ for the initial-student source probability conditional on $g$, the behavior history distribution is a mixture of complete prefix densities:
\begin{equation}
p_{\mathrm{mix},t}(h_t)
=\eta_g d_{\theta_0,t}(h_t)
+(1-\eta_g)d_{T,t}^{(j)}(h_t).
\tag{7}
\end{equation}
When conditioning preserves the source proportions, $\eta_g=\eta$. Define
\begin{equation*}
A_\nu(t)=\sum_{u<t}
\log\pi_\nu^{\mathrm{exec}}(a_u^{\mathrm{exec}}\mid h_u).
\end{equation*}
Cancellation of the shared initial-state and environment-transition factors gives
\begin{equation}
\rho_t(h_t)
=\frac{d_{\theta,t}(h_t)}{p_{\mathrm{mix},t}(h_t)}
=\frac{\exp A_\theta(t)}
{\eta_g\exp A_{\theta_0}(t)+(1-\eta_g)\exp A_T^{(j)}(t)}.
\tag{8}
\end{equation}
The denominator is a mixture of prefix products, rather than a product of per-decision mixtures. If $d_{\theta,t}$ is absolutely continuous with respect to $p_{\mathrm{mix},t}$, then for any $f$ integrable under $d_{\theta,t}$,
\begin{equation}
\mathbb E_{h_t\sim d_{\theta,t}}[f(h_t)]
=\mathbb E_{h_t\sim p_{\mathrm{mix},t}}[\rho_t(h_t)f(h_t)].
\tag{9}
\end{equation}
For fixed student parameters, $f$ may be the distillation loss averaged over fresh student denoising queries conditional on $h_t$. Only decisions preceding $t$ enter the ratio because $h_t$ is available before the current action is generated. At $t=0$, the empty prefix has ratio one under the shared initial distribution.

\subsection{Prefix Compatibility Scores}

Equation~\eqref{eq:path_score} defines a path-compatibility score $\varphi_\nu(u)$ from stored denoising transitions. This score is not assumed to be an exact executed-action log likelihood. Let $t_{\tau,k}$ denote the original decision index of the $k$-th sampled decision point in trajectory $\tau$. For $i=(\tau,k)$, the prefix score is
\begin{equation*}
\Phi_\nu(i)
=\sum_{r<k}\varphi_\nu(t_{\tau,r}).
\end{equation*}
These prefix scores are used in Equation~\eqref{eq:cum_score}, with reference-mixture coefficient $\eta$. Only preceding sampled decision points contribute, and the sum is zero for the first sampled point. Both fixed references evaluate every sampled point regardless of its collection source.

\subsection{Sampling Measure, Normalization, and Stable Computation}

We use equally spaced temporal subsampling with $n=16$ and average the loss over the sampled decision points of each trajectory. Let $\mathcal B_\tau$ index these points and let $q(\tau)$ denote the trajectory-sampling probability, including source selection. The induced sampling probability of $i=(\tau,k)$ is
\begin{equation*}
p(i)=\frac{q(\tau)}{|\mathcal B_\tau|}.
\end{equation*}

Let $I_g$ index the sampled decision points belonging to collection condition $g$. For conditions with positive probability, define
\begin{equation*}
p(g)=\sum_{i\in I_g}p(i),
\qquad
p_{i\mid g}=\frac{p(i)}{p(g)}.
\end{equation*}
All normalization uses this sampling measure, including its source-selection probabilities. With $z_i=\exp S_i$, normalization, clipping, and renormalization give
\begin{equation}
\begin{aligned}
\mu_g&=\sum_{i'\in I_g}p_{i'\mid g}z_{i'},
&r_i&=z_i/\mu_g,\\
\widetilde w_i&=\min\{r_i,C\},
&w_i&=\frac{\widetilde w_i}
{\sum_{i'\in I_g}p_{i'\mid g}\widetilde w_{i'}},
\qquad i\in I_g.
\end{aligned}
\tag{10}
\label{eq:weights}
\end{equation}
Here $C>0$ and $\lambda\geq0$. The final weights satisfy
\begin{equation*}
\sum_{i\in I_g}p_{i\mid g}w_i=1.
\end{equation*}
The cap applies before the second normalization, so the final $w_i$ need not be bounded by $C$. Setting $\lambda=0$ gives $w_i=1$ for all sampled decision points. Student-only and teacher-only references correspond to $\eta=1$ and $\eta=0$, respectively.

Conditional on the cached weights, define the reweighted empirical distribution
\begin{equation*}
p_w(i\mid g)=p_{i\mid g}w_i.
\end{equation*}
Equation~(10) ensures that $p_w(\cdot\mid g)$ is a probability distribution. Let $\ell_\theta(i)$ denote the per-decision regression loss averaged over fresh student queries conditional on the stored history. The expected replay loss can then be written as
\begin{equation*}
\mathbb E_{\tau\sim q}\left[
\frac{1}{|\mathcal B_\tau|}\sum_{i\in\mathcal B_\tau}w_i\ell_\theta(i)
\right]
=\sum_g p(g)\,
\mathbb E_{i\sim p_w(\cdot\mid g)}[\ell_\theta(i)].
\end{equation*}
Thus Equation~\eqref{eq:final_loss} defines regression under a reweighted empirical distribution over sampled decision points while preserving the sampling probabilities of the collection conditions. Because the sampled indices depend on the available trajectory length and the cumulative scores omit unsampled decisions, Equation~(8) does not directly give the density ratio for this replay measure.

For numerical stability, the reference-mixture term is evaluated with log-sum-exp:
\begin{equation}
\log\!\left(\eta e^{\Phi_{\theta_0}}+(1-\eta)e^{\Phi_T}\right)
=\operatorname{LSE}\!\left(
\log\eta+\Phi_{\theta_0},
\log(1-\eta)+\Phi_T\right),
\tag{11}
\end{equation}
where $\Phi$ abbreviates the corresponding prefix score $\Phi_\nu(i)$. At a single-source endpoint, only the corresponding term is retained. Before exponentiation within condition $g$, subtract $\max_{i\in I_g}S_i$; this common factor cancels in Equation~(10). Fixed-reference scores are cached once, while snapshot scores and weights are refreshed every $R$ updates.

\subsection{Scope of the Distillation Update}

At update $a$, let $\theta_a$ denote the student parameters used to generate fresh queries, and let $p_w^{(a)}$ denote the empirical distribution induced by the cached weights. Holding the query-generating policy and cached weights fixed, define the supervised replay surrogate
\begin{equation*}
\begin{aligned}
\mathcal R_a(\vartheta)
={}&\sum_g p(g)\,
\mathbb E_{i\sim p_w^{(a)}(\cdot\mid g)}
\mathbb E_{\substack{s\sim p_s\\\mathbf{x}\sim q_{\theta_a}(\cdot\mid h_i,s)}}
\left[
\left\|\mathbf{v}_\vartheta(\mathbf{x},s\mid h_i)
-\mathbf{v}_T^{(j_i)}(\mathbf{x},s\mid h_i)\right\|_2^2
\right].
\end{aligned}
\end{equation*}
With the query-generating student and cached weights fixed at update $a$, the detached loss in Equation~\eqref{eq:final_loss} yields a stochastic gradient of $\mathcal R_a$ evaluated at $\vartheta=\theta_a$. Gradients propagate only through the student velocity prediction. The ideal history-ratio identity motivates the prefix-score structure, while the implemented weights define the empirical replay objective in Section~A.3. These weights serve as compatibility-based replay proxies and are not assumed to be exact online importance ratios. The resulting update combines fresh student-generated denoising queries with reweighted stored histories, without additional environment interaction during distillation.

\section{Distillation Procedure}
\label{app:recipe}
The trajectory-based objective in Equation~\eqref{eq:final_loss} gives the following distillation procedure, alternating between history-weight refreshes and student updates.
\begin{enumerate}
\item \textbf{Collect the pool.} Initialize the student and task-specific teachers as described in Section~\ref{sec:exp_impl}. Collect trajectories with the initial student and each task teacher, retaining ordered histories, denoising paths, executed action lengths, and task and collection-condition labels.
\item \textbf{Cache reference scores.} Freeze the teachers and initial-student reference. Evaluate both references on every stored path using Equation~\eqref{eq:path_score} and cache the prefix scores accumulated over preceding sampled decision points.
\item \textbf{Refresh history weights.} At initialization and every $R$ updates, copy the student to the scoring snapshot. Recompute its prefix scores and obtain history weights from Equations~\eqref{eq:cum_score} and~\eqref{eq:weights}.
\item \textbf{Generate teacher queries.} Sample a trajectory from the fixed pool. At each sampled decision point, use its stored history to generate a fresh denoising path with the current student, select a low-noise state, and query the corresponding task teacher at the same state and noise level.
\item \textbf{Update the student.} Detach query states, teacher targets, and history weights. Average the weighted action-velocity losses over the sampled decision points of the trajectory and update the student. Repeat until the optimization budget is reached.
\end{enumerate}
All denoising queries and weight refreshes use stored histories. Newly sampled actions are not executed in the environment, and stored behavior actions are used for path scoring rather than as regression targets.

\paragraph{Loss reduction.}
For an action-velocity output with $D_a$ scalar coordinates, the mean squared error is $\|\mathbf v_\theta-\mathbf v_T\|_2^2/D_a$. With the action-loss coefficient of 10, the optimized loss is therefore $(10/D_a)\widehat{\mathcal L}_{\mathrm{WAM\text{-}OPD}}$. The executed-coordinate projection $P_u$ is used only in the path score, not in this action-field regression.

\section{Evaluation Protocol and Interaction Cost}
\label{app:evaluation}
\paragraph{Simulation.}
For task $j$, success rate is $100N_j^{\mathrm{succ}}/N_j$, where $N_j^{\mathrm{succ}}$ and $N_j$ denote successful and total trials under the evaluation setting in Section~\ref{sec:exp_setup}. Target and retention averages assign equal weight to their two and six constituent tasks, respectively. Improvements are measured in percentage points before rounding. All compared policies share evaluation scenes, task instructions, and inference-noise seeds. Ablation results are computed over 200 cases per task.

\paragraph{Real-world evaluation.}
Each method is evaluated on 50 paired layouts per task, disjoint from the training layouts. Methods are tested on the same layouts in interleaved order. A pick-and-place trial succeeds when the object reaches the designated region and remains stable after release; a button-press trial succeeds when the button is activated. Time limits, target regions, and stability criteria are fixed across methods.

\paragraph{Interaction cost.}
Environment interaction comprises teacher-data collection, fixed-pool collection, and closed-loop evaluation. Distillation itself requires no additional environment interaction: both student queries and teacher supervision are computed from stored histories. This accounting separates the one-time collection cost from the interaction-free optimization stage; teacher inference and history-weight refreshes still incur computation.

\section{Additional Experimental Details}
\label{app:configuration}

\subsection{Simulation teachers and trajectory pool}
\label{app:server_pool}
The Stapler and Microwave teachers are obtained after 32,000 and 16,000 SFT steps, respectively. The Microwave demonstration dataset contains 2,492 episodes, comprising 492 clean and 2,000 randomized demonstrations, with a 1\% validation split. Teacher-training demonstrations are separate from the rollout pool used for distillation.

Table~\ref{tab:server_pool} gives the fixed-pool composition.
\begin{table}[ht]
\centering
\caption{Composition of the fixed simulation trajectory pool.}
\label{tab:server_pool}
\small
\begin{tabular}{@{}llrrr@{}}
\toprule
Task & Source & Clean & Randomized & Decision records \\
\midrule
Stapler & Initial student & 8 & 32 & 443 \\
Stapler & Teacher & 4 & 16 & 166 \\
Microwave & Initial student & 8 & 32 & 1,802 \\
Microwave & Teacher & 4 & 16 & 780 \\
\midrule
Total & & 24 & 96 & 3,191 \\
\bottomrule
\end{tabular}
\end{table}

\subsection{Distillation hyperparameters}
Table~\ref{tab:configuration} summarizes the optimization, path-scoring, and prefix-weighting hyperparameters. Student queries use the current model parameters. Reference scores are cached, and the student snapshot and history weights are refreshed at updates 0, 24, 48, and 72. Path residuals are computed in FP32 with FP64 reductions.
\begin{table}[ht]
\centering
\caption{Hyperparameters for simulation distillation.}
\label{tab:configuration}
\small
\begin{tabular}{@{}p{0.56\linewidth}p{0.36\linewidth}@{}}
\toprule
Parameter & Value \\
\midrule
Trainable parameters & All student MoT parameters \\
Optimizer & AdamW \\
Learning rate & $10^{-6}$ \\
Adam coefficients $(\beta_1,\beta_2)$ & $(0.9,0.95)$ \\
Weight decay & 0 \\
Maximum gradient norm & 1.0 \\
Training precision & BF16 \\
Optimization steps & 96 \\
Temporal subsampling & Equally spaced, $n=16$ \\
Action / video loss coefficients & 10 / 0 \\
Reweighting strength $\lambda$ & 0.1 \\
Scoring scale $\kappa$ & 0.025 \\
Minimum scoring scale $\sigma_{\min}$ & $10^{-4}$ \\
Clipping threshold $C$ & 2 \\
Weight refresh interval $R$ & 24 \\
Denoising steps & 10 \\
Prediction / execution horizon & 32 / up to 24 \\
\bottomrule
\end{tabular}
\end{table}
The clipping threshold applies before the second normalization in Equation~\eqref{eq:weights}; the final normalized weights may exceed $C$.

\paragraph{Baselines.}
The task-specific SFT teachers, $\pi$RL~\citep{chen2025pirl}, and STEAM~\citep{liu2026steam} share the Fast-WAM initialization used in the simulation experiments. All policies use the inference horizon and denoising schedule in Table~\ref{tab:configuration}.

\subsection{Real-robot observation and action interface}
\label{app:real_records}
The platform consists of two UR5e arms and three RGB cameras. The head camera captures the workspace, while the two wrist cameras provide local views of the grippers and manipulated objects.

The action vector has 14 dimensions: six target joint positions and one gripper command for the left arm, followed by the corresponding seven values for the right arm. Actions specify absolute joint targets, with gripper values scaled by $1/1000$. Proprioception contains measured joint positions and gripper states in the same order. Shared dataset statistics are used to normalize observations and actions across tasks.

\section{Limitations}
WAM-OPD's performance depends on the coverage of the fixed trajectory pool and the accuracy of the history-weight approximation. Broader coverage of task-relevant histories and more accurate history weighting are directions for future work.

\end{document}